\documentclass[conference,a4paper]{IEEEtran}
\IEEEoverridecommandlockouts

\usepackage{cite}
\usepackage{amsmath,amssymb,amsfonts}
\usepackage{algorithmic}
\usepackage{graphicx}
\usepackage{textcomp}
\usepackage{xcolor}

\usepackage[table]{xcolor}
\usepackage{subcaption}
\usepackage{tabularx}
\usepackage{booktabs}
\usepackage{makecell}
\usepackage{xurl}
\usepackage{xspace}
\usepackage{multirow}

\newcommand{\ours}{DGVC\xspace}

\def\BibTeX{{\rm B\kern-.05em{\sc i\kern-.025em b}\kern-.08em
    T\kern-.1667em\lower.7ex\hbox{E}\kern-.125emX}}
\begin{document}

\title{Dense-Prior-Guided Generative Video Compression}

\author{
\IEEEauthorblockN{
Ding Ding$^{1\dagger}$,
Daowen Li$^{1\dagger}$,
Yixin Gao$^{1}$,
Ruixiao Dong$^{1,2}$,
Kai Li$^{1*}$,
Ying Chen$^{1*}$,
Li Li$^{2}$
}
\IEEEauthorblockA{$^{1}$Alibaba Group\\
\{liangjie.dd, lidaowen.ldw, gaoyixin.gyx, dongruixiao.drx, kaishi.lk, YingChen\}@alibaba-inc.com}
\IEEEauthorblockA{$^{2}$Institute of Artificial Intelligence, Hefei Comprehensive National Science Center\\
\{lili90th\}@gmail.com}
}

\maketitle

\begingroup
\renewcommand\thefootnote{}
\footnotetext{$^{\dagger}$ Equal contribution.}
\footnotetext{$^{*}$ Corresponding author.}
\endgroup
\setcounter{footnote}{0}

\begin{abstract}

Diffusion-based generative video compression offers a promising paradigm for low-bitrate reconstruction, but existing keyframe-based controllable approaches rely on sparse priors for non-keyframes, which often lack sufficient low-level details for faithful reconstruction and require non-trivial training to adapt generative models to specialized controls. We propose \textbf{D}ense-Prior-Guided \textbf{G}enerative \textbf{V}ideo \textbf{C}ompression (\textbf{DGVC}), a training-free framework that uses codec-compressed luminance components of non-keyframes as dense priors to guide reconstruction. Unlike sparse controls, the dense luminance priors are compatible with modern fidelity-oriented video codecs and remain sufficiently faithful after compression, eliminating the need for prior-specific codec training and making additional generative-model adaptation unnecessary in practice. Moreover, luminance priors preserve richer structural and textural information, thereby providing more reliable guidance for controllable video generation. DGVC further introduces a color-distance-guided keyframe selection strategy to capture color variations for chrominance recovery. Extensive experiments show that DGVC consistently surpasses prior controllable compression methods in both signal fidelity and perceptual quality, while remaining competitive with representative post-enhancement approaches.

\end{abstract}

\begin{IEEEkeywords}
Generative Video Compression, Controllable Video Generation, Perceptual Video Compression
\end{IEEEkeywords}

\begin{figure*}[t]
\centering
    \includegraphics[width=\textwidth]{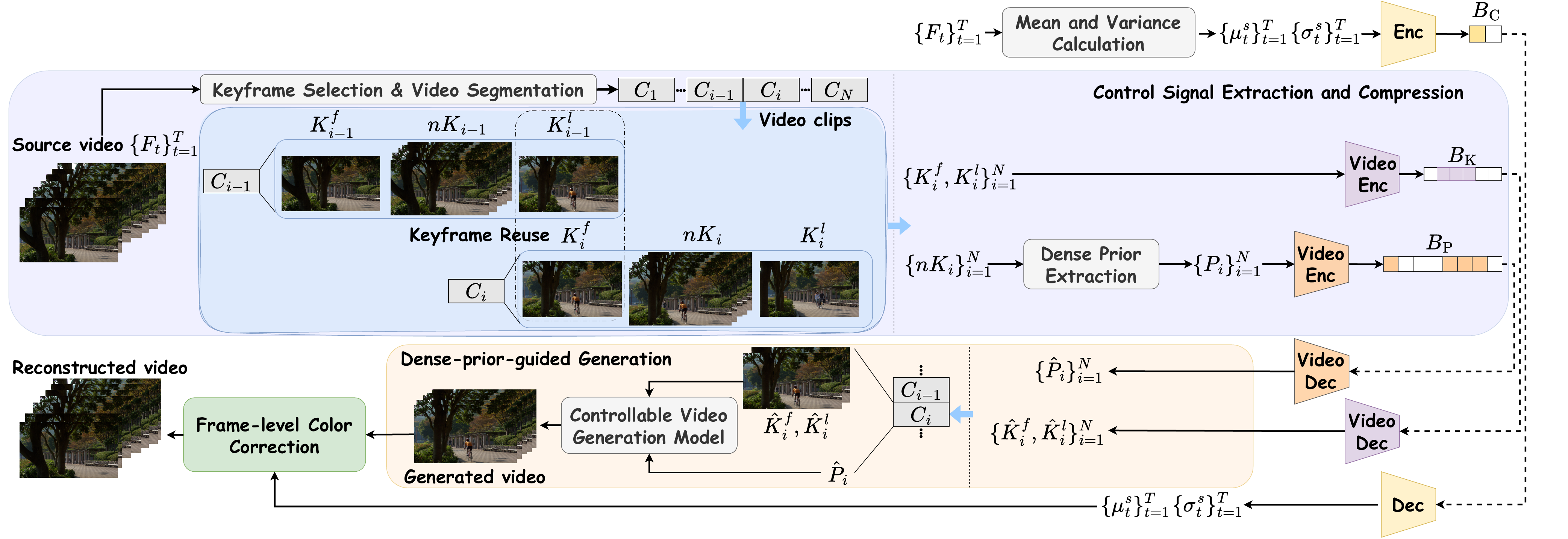}
    \vspace{-22.5pt}
    \caption{Framework of the \ours, which first selects keyframes and extracts luminance priors at the encoder, compresses them with a standard codec, and performs clip-wise controllable generation and frame-wise color correction at the decoder.}
    \label{fig:framework}
    \vspace{-15pt}
\end{figure*}

\section{Introduction}
\label{sec:intro}

Video data dominates modern Internet traffic, making efficient video compression a long-standing research problem. Beyond fidelity-oriented codecs\cite{wiegand2003overview,sullivan2012overview,bross2021overview,li2024neural} optimized for pixel-wise distortion, recent progress in generative models, particularly diffusion models \cite{liu2024sora,yang2025cogvideox,wan2025wan}, has enabled \emph{generative video compression} methods to improve perceptual quality based on generative priors. Existing work mainly falls into two paradigms: \emph{post-enhancement}, which enhances codec reconstructions with generative models \cite{li2024extreme,liu20242,ma2025diffusion}, and \emph{keyframe-based controllable compression}, which encodes control signals decomposed from videos to guide generation in decoding \cite{wan2025m3,wang2026t,11464112,CVGEVC}. A notable advantage of keyframe-based controllable compression is that it shifts more of the reconstruction burden from source coding to generation, thereby reducing bitrate.

Existing keyframe-based controllable compression methods encode keyframes with sparse priors for non-keyframes, such as text, trajectories, poses, or segmentation cues, to guide video generation \cite{wan2025m3,wang2026t,11464112,CVGEVC}. However, these sparse controls capture high-level semantics or coarse motion, limiting faithful reconstruction. Moreover, because they are task-specific and unsupported by standard video codecs, they often require specialized coding schemes or separately trained codecs. Existing methods also typically fine-tune controllable generation models to adapt to reconstructed priors, especially when such priors are not natively supported. This raises a fundamental question: \emph{can controllable generative video compression exploit a dense, codec-friendly control signal that preserves visual detail without additional training?}

In this paper, we propose \textbf{Dense-Prior-Guided Generative Video Compression (DGVC)}, a training-free framework for keyframe-based controllable compression. Our key insight is that the luminance component of non-keyframes serves as an effective dense prior, preserving rich structural and textural information while being efficiently compressed by advanced video codecs. Moreover, pretrained controllable video generation models \cite{jiang2025vace} can directly use luminance priors as condition images during decoding. When the compressed luminance prior retains high fidelity, DGVC guides reconstruction without additional model adaptation. To support chrominance recovery, we introduce a color-distance-guided keyframe selection strategy that places keyframes based on object-level color variation. Extensive experiments show that DGVC outperforms prior keyframe-based controllable compression methods in signal fidelity and perceptual quality, while remaining competitive with representative post-enhancement methods.

Our main contributions can be concluded as follows:
\begin{itemize}
    \item We propose a training-free framework, namely \textbf{DGVC}, for keyframe-based controllable generative video compression, which leverages codec-compressed luminance components as dense priors for reconstruction.
    \item We design a \textbf{color-distance-guided keyframe selection} strategy that facilitates chrominance recovery by adaptively placing keyframes to capture color variation.
    \item Experiments show that DGVC yields state-of-the-art performance within the controllable compression paradigm.
\end{itemize}

\section{Related Work}

\subsection{Generative Video Compression}

Early work on generative video compression mainly improved perceptual quality using generative adversarial network (GAN) \cite{mentzer2022neural,lu2019dvc,yang2022perceptual}. More recent diffusion-based methods mainly fall into two paradigms: \emph{post-enhancement} \cite{li2024extreme,liu20242,ma2025diffusion} and \emph{keyframe-based controllable compression} \cite{wan2025m3,wang2026t,11464112,CVGEVC}. Existing controllable compression methods typically encode keyframes together with sparse priors for non-keyframes, such as text \cite{wan2025m3}, trajectories \cite{wang2026t}, camera poses and segmentation cues \cite{11464112}, or multimodal controls \cite{CVGEVC}. Most methods rely on specialized codecs to compress these control signals as well as task-specific fine-tuning of controllable generation models. In contrast, our method explores a dense and codec-friendly prior for keyframe-based controllable compression without training.

\subsection{Controllable Video Generation Models}
Recent video diffusion models \cite{liu2024sora,yang2025cogvideox,wan2025wan,ma2025controllable} have enabled controllable video generation from multimodal inputs, such as text, reference frames, and motion cues. Among them, VACE \cite{jiang2025vace} provides a versatile backbone for controllable video generation and supports diverse control signals, including the keyframes and luminance priors used in \ours. In addition, pretrained on large-scale datasets, VACE shows robustness to control inputs with compression artifacts, which further supports the training-free design of \ours.

\section{Proposed Method}
\subsection{Overview}

As illustrated in Fig.~\ref{fig:framework}, \ours compresses a source video by transmitting two types of control signals, namely selected keyframes and dense priors formed by the luminance components of non-keyframes. Both signals are encoded by a modern fidelity-oriented video codec and decoded at the receiver, respectively. The decoded controls are then fed into a pretrained controllable video generation model for clip-wise reconstruction, without task-specific fine-tuning.

\textbf{Control signal extraction and compression.}
Given an input video $\{F_t\}_{t=1}^T$, \ours first selects representative keyframes to partition it into $N$ clips $\{C_i\}_{i=1}^N$. For each clip $C_i$, the first and last keyframes, denoted as $K_i^f$ and $K_i^l$, serve as reference frames for generation. To maintain temporal continuity, the last keyframe of $C_{i-1}$ is reused as the first keyframe of $C_i$. The remaining frames are represented by a dense prior $P_i$ composed of their luminance components. Unlike sparse controls, luminance priors preserve richer structural and textural cues and can be efficiently compressed by standard video codecs. DGVC encodes the keyframes and dense priors into bitstreams $B_K$ and $B_P$, respectively.

\textbf{Dense-prior-guided generation.}
At the decoder, the keyframes and dense priors are recovered from the bitstreams. A pretrained controllable video generation model, VACE~\cite{jiang2025vace}, takes the reconstructed first and last keyframes $\{\hat{K}_i^f,\hat{K}_i^l\}$ together with dense prior $\hat{P}_i$ as input, and reconstructs each clip in a first-last-frames-to-video manner. The reconstructed clips are subsequently concatenated to form the output video.

\textbf{Color correction.}
Since the generated frames may exhibit color shifts, \ours applies a lightweight frame-level correction after generation. Specifically, each reconstructed frame is modulated using the mean and variance transmitted from its corresponding source frame,
\begin{equation}
F_t^c = (F_t^r - \mu_t^r)\cdot \frac{\sigma_t^s}{\sigma_t^r} + \mu_t^s,
\label{eq:color_correction}
\end{equation}
where $F_t^r$ and $F_t^c$ denote the reconstructed and corrected frames, $\mu_t^r$ and $\sigma_t^r$ are the mean and variance of $F_t^r$, and $\mu_t^s$ and $\sigma_t^s$ are those computed from the corresponding source frame. The source-frame statistics $\mu_t^s$ and $\sigma_t^s$ are losslessly encoded into an auxiliary bitstream $B_C$.

\begin{figure*}[t]
  \centering
  \begin{subfigure}{0.25\linewidth}
    \includegraphics[width=\textwidth]{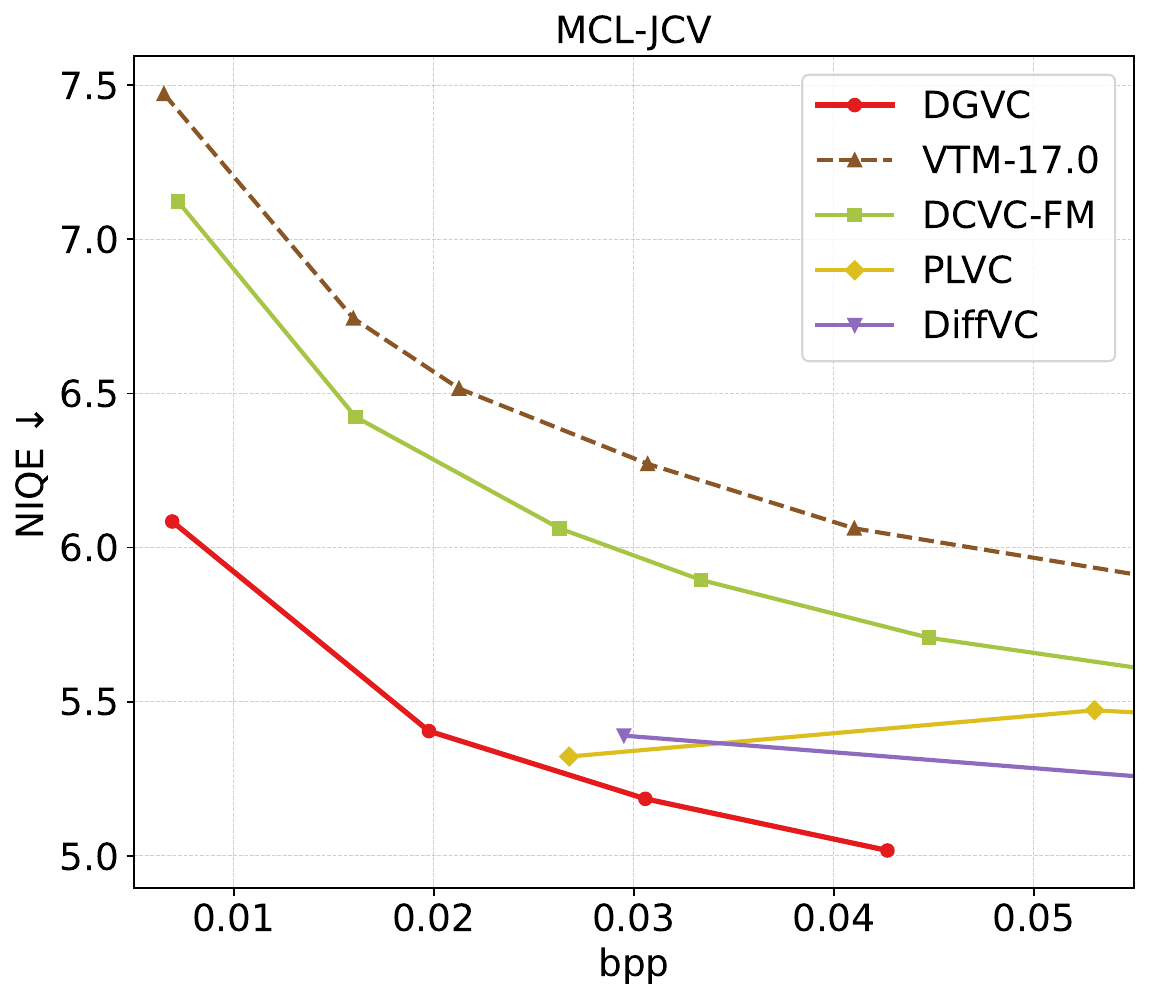}
    % \caption{PSNR}
    \label{fig:MCL-JCV_niqe}
  \end{subfigure}
    \hfill
    \hspace{-10pt}
  \begin{subfigure}{0.25\linewidth}
    \includegraphics[width=\textwidth]{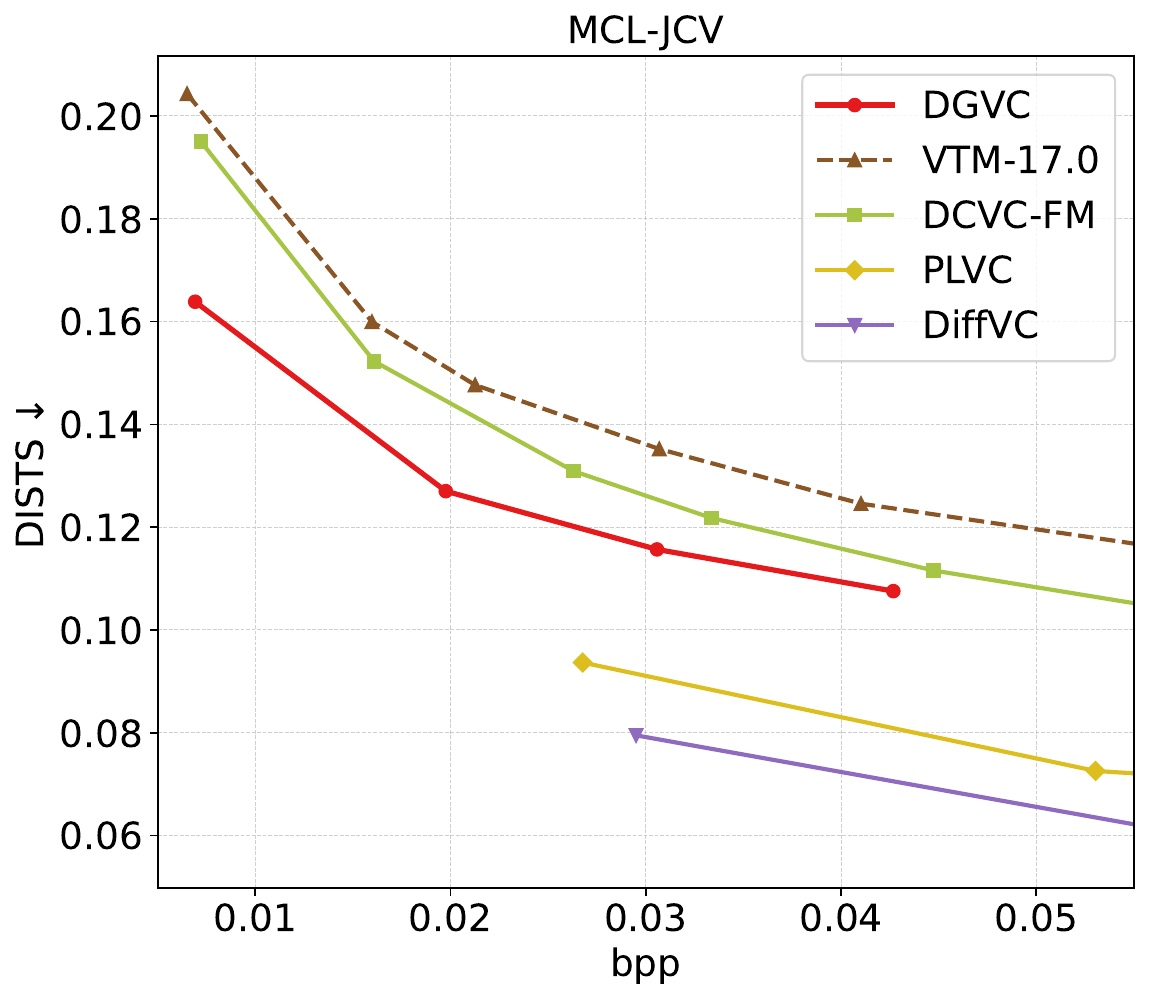}
    % \caption{VFID}
    \label{fig:MCL-JCV_dist}
  \end{subfigure}
      \hfill
      \hspace{-10pt}
  \begin{subfigure}{0.25\linewidth}
    \includegraphics[width=\textwidth]{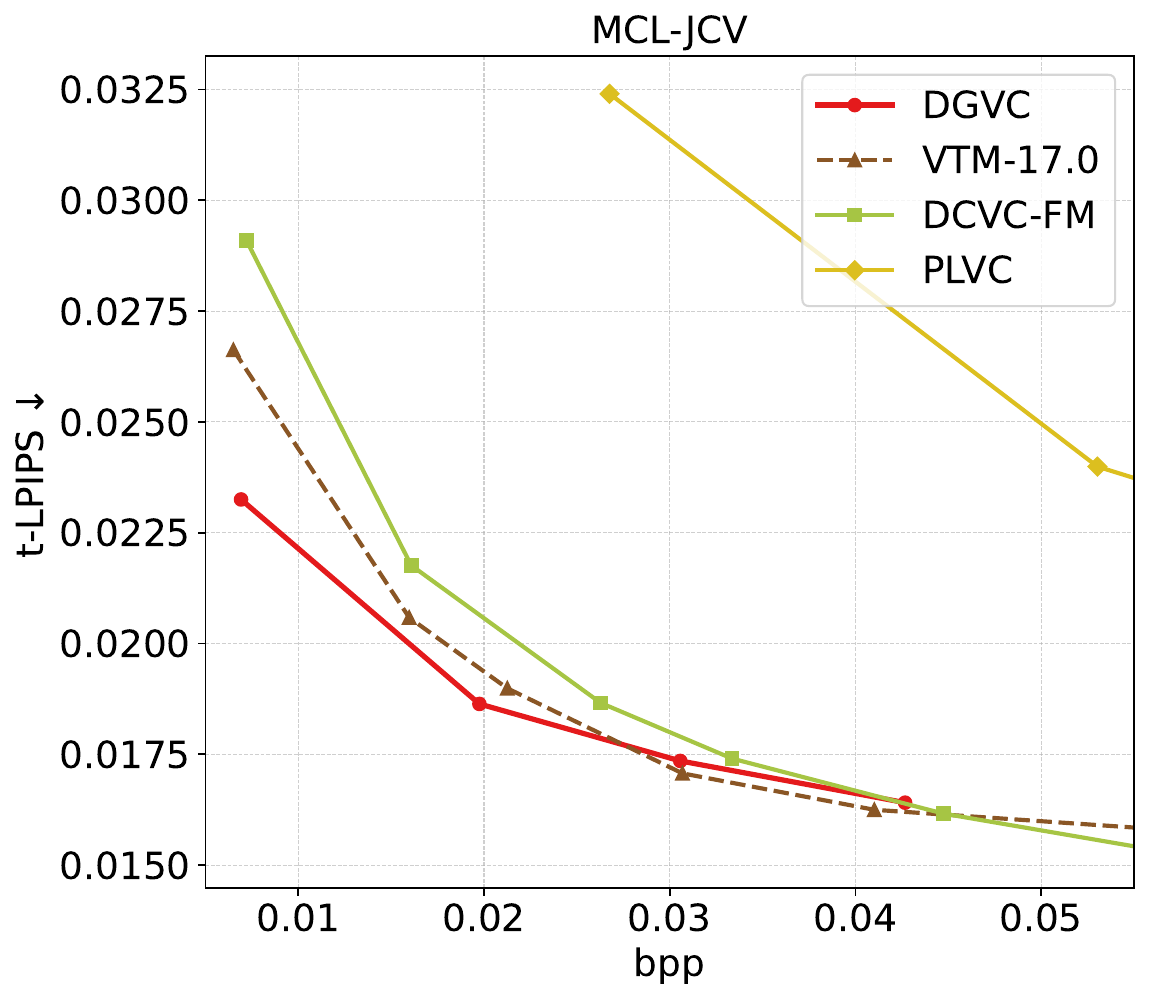}
    % \caption{VFID}
    \label{fig:MCL-JCV_tlpips}
  \end{subfigure}
      \hfill
      \hspace{-10pt}
  \begin{subfigure}{0.25\linewidth}
    \includegraphics[width=\textwidth]{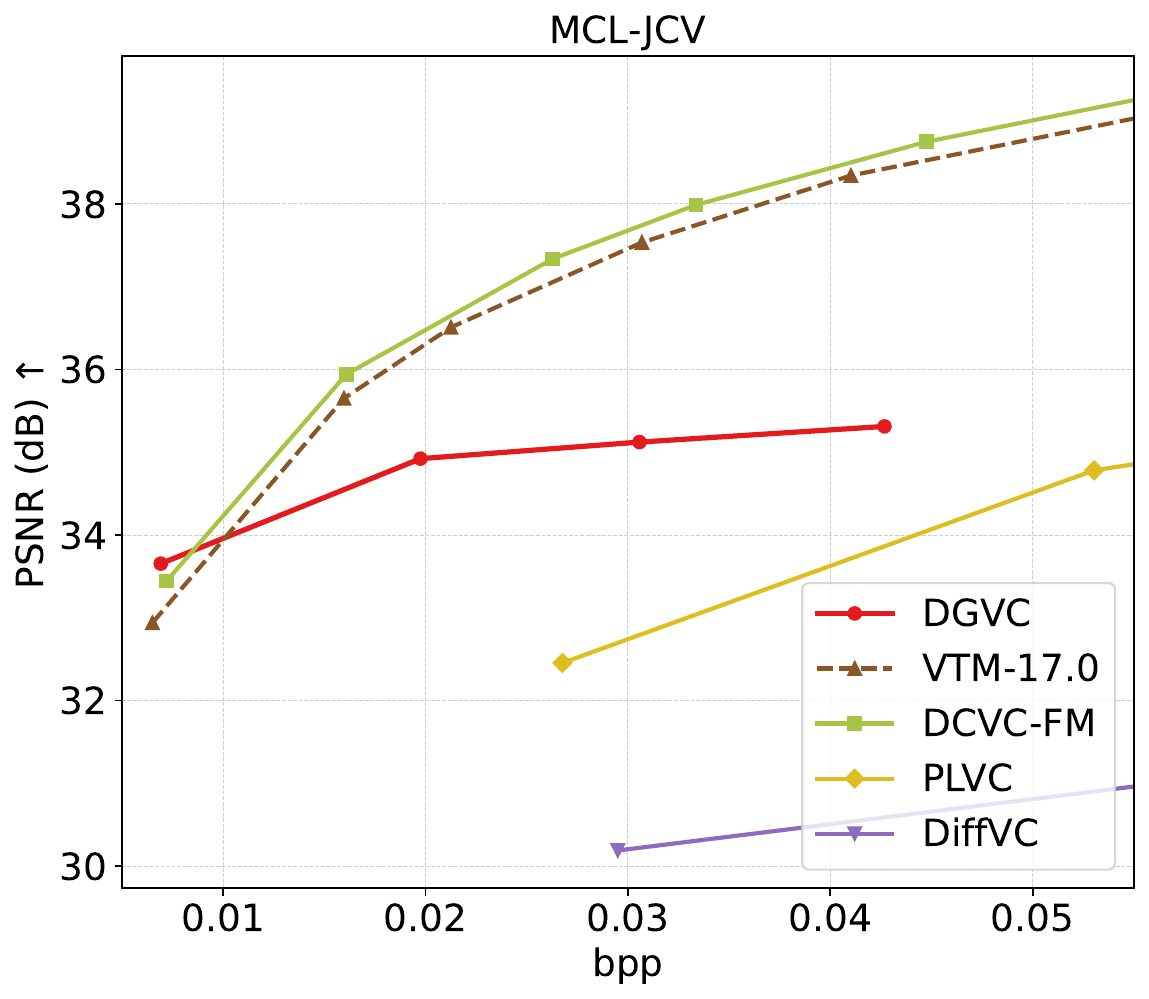}
    % \caption{VFID}
    \label{fig:MCL-JCV_psnr}
  \end{subfigure}

  \vspace{-20pt}
  \caption{Rate and perception/fidelity curves on MCL-JCV.
  }
  \label{fig:RD curve comparison}
  \vspace{-5pt}
\end{figure*}

\begin{table*}[t]
    \begin{center}
    \caption{\label{tab:comparison}
    BD-rate ($\%$)$\downarrow$ / BD-metric$\uparrow$ on MCL-JCV, HEVC B, and UVG, using VTM-17.0 as the anchor. The best performances are highlighted in \textbf{bold}. ``N/A'' indicates no quality overlap for BD-rate calculation. ``-'' denotes unavailable data.
    }
    \vspace{-5pt}
    \renewcommand{\arraystretch}{1.25}
    \renewcommand{\tabcolsep}{0.8mm}
    \scalebox{0.82}{
    \begin{tabular}{l|c|c|c|c||c|c|c|c||c|c|c|c}
    \hline
    \cline{1-13}
    & \multicolumn{4}{c||}{MCL-JCV} & \multicolumn{4}{c||}{HEVC B} & \multicolumn{4}{c}{UVG} \\
    \cline{1-13}
    Method &
    NIQE & DISTS & t-LPIPS & PSNR (dB) &
    NIQE & DISTS & t-LPIPS & PSNR (dB) &
    NIQE & DISTS & t-LPIPS & PSNR (dB) \\
    \cline{1-13}
    DCVC-FM
      & -28.4 / 0.28 & -11.3 / 0.0057 & 30.0 / -0.0022 & \textbf{-6.21} / \textbf{0.19}
      & -31.7 / 0.28 & -1.70 / 0.0006 & -7.40 / 0.0004 & \textbf{-13.9} / \textbf{0.38}
      & -33.2 / 0.25 & -13.1 / 0.0060 & 15.7 / -0.0014 & \textbf{-18.7} / \textbf{0.52} \\
    \hline
    PLVC
      & N/A / 0.65 & N/A / 0.0454 & 477 / -0.0108 & 353 / -4.39
      & N/A / 0.43 & \textbf{N/A} / \textbf{0.0436} & 256 / -0.0063 & 356 / -4.02
      & N/A / 1.01 & N/A / 0.0486 & 276 / -0.0074 & 473 / -4.27 \\
    \hline
    DiffVC
      & N/A / 0.75 & \textbf{N/A} / \textbf{0.0554} & - / - & 1596 / -7.74
      & - / - & - / - & - / - & - / -
      & N/A / 0.85 & \textbf{N/A} / \textbf{0.0545} & - / - & 2336 / -7.61 \\
    \hline
    \rowcolor[gray]{0.9}
    DGVC
      & \textbf{-83.4} / \textbf{1.19} & -50.1 / 0.0261 & \textbf{-18.6} / \textbf{0.0011} & 20.1 / -1.15
      & \textbf{-78.4} / \textbf{1.07} & -41.1 / 0.0186 & \textbf{-12.8} / \textbf{0.0007} & 35.0 / -1.25
      & \textbf{N/A} / \textbf{1.79} & -48.9 / 0.0284 & \textbf{-13.1} / \textbf{0.0010} & 46.2 / -1.14 \\
    \hline
    \hline
    \multicolumn{13}{l}{\textit{The following results are reported on source sequences resized to \textbf{320$\times$512}, in accordance with the configuration proposed in the T-GVC paper.}} \\
    \hline
    \hline
    T-GVC
      & -49.4 / 0.51 & -28.5 / 0.0096 & N/A / -0.0287 & N/A / -8.72
      & -57.6 / 0.64 & -25.0 / 0.0082 & N/A / -0.0169 & N/A / -8.11
      & -24.7 / 0.27 & -8.82 / 0.0047 & N/A / -0.0133 & N/A / -8.48 \\
    \hline
    \rowcolor[gray]{0.9}
    DGVC
      & \textbf{-71.4} / \textbf{0.93} & \textbf{-46.4} / \textbf{0.0190} & \textbf{-8.05} / \textbf{0.0002} & \textbf{88.2} / \textbf{-3.40}
      & \textbf{-74.5} / \textbf{0.99} & \textbf{-49.3} / \textbf{0.0177} & \textbf{-9.03} / \textbf{0.0006} & \textbf{65.1} / \textbf{-2.54}
      & \textbf{-62.6} / \textbf{0.80} & \textbf{-14.3} / \textbf{0.0065} & \textbf{-9.57} / \textbf{0.0003} & \textbf{136} / \textbf{-2.72} \\
    \hline
    \cline{1-13}
    \end{tabular}
    }
    \end{center}
    \vspace{-15pt}
\end{table*}

\begin{figure*}[t]
    \centering
    {\small
        \begin{tabularx}{\textwidth}{
          >{\centering\arraybackslash\hsize=2.15\hsize\linewidth=\hsize}X
          >{\centering\arraybackslash\hsize=0.77\hsize\linewidth=\hsize}X
          >{\centering\arraybackslash\hsize=0.77\hsize\linewidth=\hsize}X
          >{\centering\arraybackslash\hsize=0.77\hsize\linewidth=\hsize}X
          >{\centering\arraybackslash\hsize=0.77\hsize\linewidth=\hsize}X
          >{\centering\arraybackslash\hsize=0.77\hsize\linewidth=\hsize}X
        }
        & Ground Truth & VTM-17.0 & DCVC-FM & PLVC & \textbf{\ours} \\
        \end{tabularx}
    }
    \includegraphics[width=\textwidth]{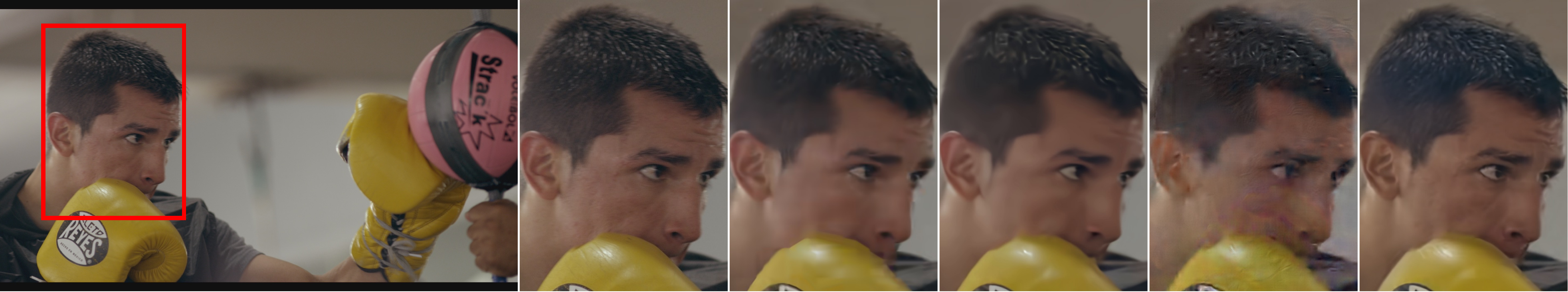}
    {\small
        \begin{tabularx}{\textwidth}{
          >{\centering\arraybackslash\hsize=2.15\hsize\linewidth=\hsize}X
          >{\centering\arraybackslash\hsize=0.77\hsize\linewidth=\hsize}X
          >{\centering\arraybackslash\hsize=0.77\hsize\linewidth=\hsize}X
          >{\centering\arraybackslash\hsize=0.77\hsize\linewidth=\hsize}X
          >{\centering\arraybackslash\hsize=0.77\hsize\linewidth=\hsize}X
          >{\centering\arraybackslash\hsize=0.77\hsize\linewidth=\hsize}X
        }
        & BPP$\downarrow$ / NIQE$\downarrow$ & 0.0159 / 6.63 & 0.0167 / 6.21 & 0.0332 / 5.67 & 0.0152 / 5.24 \\
        & DISTS$\downarrow$ & 0.1586 & 0.1542 & 0.1145 & 0.1331 \\
        \end{tabularx}
    }
    \vspace{-10pt}
    \caption{Visual comparisons with open-source baselines on the \textit{videoSRC26} sequence (1080$\times$1920) from MCL-JCV.}
    \label{fig:visualization on videoSRC26}
    \vspace{-15pt}
\end{figure*}

\subsection{Color-distance-guided Keyframe Selection}
\label{subsection:keyframe selection}

In DGVC, the chrominance of non-keyframes is inferred from the keyframes of each clip. Uniform keyframe sampling may fail to capture large object-level color changes caused by occlusion, viewpoint changes, or object rotation, which thereby degrades color fidelity during generation. Adding more keyframes can alleviate this issue but would increase bitrate cost. DGVC therefore selects keyframes adaptively according to object-level color variation, aiming to better balance coding cost and chrominance recovery. Intuitively, keyframes should cover the major color appearances of objects over time so that the generator can infer non-keyframe chrominance from more representative references.

Given a source video $\{F_t\}_{t=1}^T$, we segment objects $\{O_m\}_{m=1}^M$ using an off-the-shelf video object segmentation model \cite{entitysam}, where $M$ is the number of segmented objects. For each object $O_m$ in frame $F_t$, its pixels are converted into an $\ell_1$-normalized color histogram $\hat{H}_t^m$ with 16 bins per channel.

By comparing the differences in $\{\hat{H}_t^m\}_{t=1}^T$ across frames, we perform a bidirectional keyframe search. The forward and backward searches start from the first and last frames of the video, respectively, each covering half of the sequence. In the forward search, the keyframe set $\mathcal{KF}_f$ is initialized as $\{F_1\}$, and the previous keyframe index $t_\text{preK}$ is set to 1. The next keyframe is then selected from the candidate frames $\{F_{t_\text{cand}} \mid t_\text{cand} \in [t_\text{preK} + W_{\min}, t_\text{preK} + W_{\max}]\}$, where $W_{\min}$ and $W_{\max}$ are the predefined minimum and maximum keyframe intervals.

For object $O_m$, the color-distribution distance between the previous and candidate keyframes is measured by
\begin{equation}
D_m(t_{\text{preK}}, t_{\text{cand}})
= \sum_{j=1}^{16^3} \max\bigl(0,\hat{H}_{t_{\text{cand}}}^{m}[j] - \hat{H}_{t_{\text{preK}}}^{m}[j]\bigr),
\label{eq:positive_color_histogram_distance}
\end{equation}
where $\hat{H}_{t_{\text{preK}}}^{m}$ and $\hat{H}_{t_{\text{cand}}}^{m}$ denote the color histograms of $F_{t_\text{preK}}$ and $F_{t_\text{cand}}$ for object $O_m$. A larger $D_m$ indicates that $F_{t_\text{cand}}$ introduces more previously unseen colors for object $O_m$.

Across all objects, we define a candidate set $\mathcal{C}$ based on $D_m$:
\begin{equation}
    \begin{aligned}
        \mathcal{C}=& \{ t_\text{cand} \mid D_m (t_{\text{preK}}, t_{\text{cand}}) \ge \tau, \\ & t_\text{cand} \in [t_\text{preK} + W_{\min}, t_\text{preK} + W_{\max}], 1\le m \le M\},
    \end{aligned}
\end{equation}
where $\tau$ is a threshold. To aggregate the discrete candidates in $\mathcal{C}$ into a single keyframe decision, we apply kernel density estimation (KDE) \cite{KDE} to fit a density function $f(t)$ over $\mathcal{C}$. Isolated candidates are smoothed out into negligible densities, whereas temporally clustered object-level color changes accumulate into a distinct peak:
\begin{equation}
f(t) = \frac{1}{5|\mathcal{C}|}
\sum_{t_\text{cand} \in \mathcal{C}}
\frac{1}{\sqrt{2\pi}}
\exp\left(-\frac{1}{2}\left(\frac{t - t_\text{cand}}{5}\right)^2\right),
\label{eq:kde}
\end{equation}
where $t \in [t_\text{preK} + W_{\min}, t_\text{preK} + W_{\max}]$. The frame corresponding to the peak of $f(t)$ is selected as the next keyframe and added to $\mathcal{KF}_f$. Its frame index then becomes the new $t_\text{preK}$ for the subsequent search.
The backward search is performed analogously to construct $\mathcal{KF}_b$. The final keyframe set is given by $\mathcal{KF}_f \cup \mathcal{KF}_b$ for subsequent compression and generation,
which balances coding efficiency and color fidelity.

\section{Experiments}

\subsection{Experimental Protocol}

\subsubsection{Evaluation}
\

\textit{\textbf{Dataset}}. \ours is evaluated on 1080p datasets, including MCL-JCV \cite{MCL-JCV}, HEVC B \cite{sullivan2012overview}, and UVG \cite{UVG}. Following prior work \cite{yang2022perceptual,wang2026t}, we use 96 frames from each video in the RGB color space, where the original YUV420 videos are converted to RGB using the BT.709 standard \cite{li2023neural}.

\textit{\textbf{Baseline}}. \ours is compared with two types of baselines: (a) fidelity-oriented video codecs, including VTM-17.0 \cite{bross2021overview} and DCVC-FM \cite{li2024neural}; and (b) generative video codecs, including PLVC \cite{yang2022perceptual}, DiffVC \cite{ma2025diffusion}, and T-GVC \cite{wang2026t}. Notably, DiffVC and T-GVC are representative methods for the post-enhancement and controllable compression paradigms, respectively. For VTM-17.0, we adopt the configuration in \cite{li2023neural}. Since the code for DiffVC is unavailable, we report the results from the original paper. In T-GVC, all six quality points from MS-ILLM \cite{muckley2023improving} are tested for BD-rate / BD-metric calculation, rather than only two as reported in their paper. For fair comparison with T-GVC, we additionally evaluate DGVC on videos resized to 320$\times$512 under the setting of \cite{wang2026t}.

\textit{\textbf{Metric}}. PSNR is reported to measure signal fidelity. Perceptual quality is assessed by the reference-based metric DISTS \cite{DISTS} and t-LPIPS \cite{T-LPIPS}, as well as the no-reference metric NIQE \cite{NIQE}. Compression performance is evaluated using BD-rate and BD-metric \cite{yang2022perceptual}, with bitrate measured in bit per pixel (BPP). 

\subsubsection{Implementation Details}
We use VVenC \cite{VVenC} for control signal compression, with GOP size of 32 and intra period of -1. The quantization parameters (QPs) for keyframes and luminance priors are set to $(25, 27)$, $(28, 30)$, $(33, 34)$, and $(40, 41)$, where keyframes account for approximately 30\% of the total bitrate. 
In color-distance-guided keyframe selection, $\tau$ is set to 0.4, with $W_{\min}$ = 16 and $W_{\max}$ = 64.
For VACE\cite{jiang2025vace}, we use the 1.3B version with 20 denoising steps.

\subsection{Comparison Results}

\newsavebox{\subfigboxA}
\newsavebox{\subfigboxB}
\newsavebox{\subfigboxC}
\newlength{\subfigheight}
\newlength{\rotlabelwidth}
\setlength{\rotlabelwidth}{0.045\columnwidth}

\begin{figure}[t]
    \centering
    \captionsetup[subfloat]{labelformat=empty,skip=2pt}

    % ---------- Subfigure 1 ----------
    \sbox{\subfigboxA}{%
        \begin{minipage}{0.96\columnwidth}
            \centering
            {\small
                \begin{tabularx}{\linewidth}{
                  >{\centering\arraybackslash\hsize=1\hsize\linewidth=\hsize}X
                  >{\centering\arraybackslash\hsize=1\hsize\linewidth=\hsize}X
                  >{\centering\arraybackslash\hsize=1\hsize\linewidth=\hsize}X
                  >{\centering\arraybackslash\hsize=1\hsize\linewidth=\hsize}X
                }
                $t=20$ & $t=30$ & $t=40$ & $t=50$ \\
                \end{tabularx}
            }
            \includegraphics[width=\linewidth]{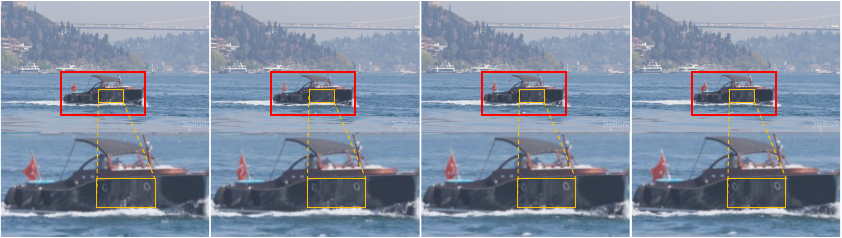}
        \end{minipage}%
    }
    \setlength{\subfigheight}{\dimexpr\ht\subfigboxA+\dp\subfigboxA\relax}

    \subfloat[Sequence-level BPP$\downarrow$ / NIQE$\downarrow$ / DISTS$\downarrow$ / t-LPIPS$\downarrow$]{%
        \makebox[\columnwidth][c]{%
            \parbox[c][\subfigheight][c]{\rotlabelwidth}{\centering\rotatebox{90}{\small Ground Truth}}%
            \hspace{1pt}%
            \usebox{\subfigboxA}%
        }%
    }

    % \vspace{2pt}

    % ---------- Subfigure 2 ----------
    \sbox{\subfigboxB}{%
        \begin{minipage}{0.96\columnwidth}
            \centering
            \includegraphics[width=\linewidth]{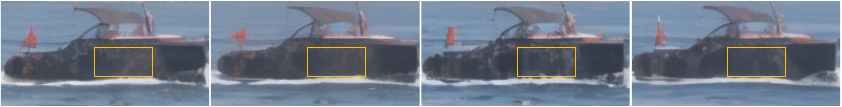}
        \end{minipage}%
    }
    \setlength{\subfigheight}{\dimexpr\ht\subfigboxB+\dp\subfigboxB\relax}

    \subfloat[0.00437 /  5.58 / 0.1699 / 0.0151]{%
        \makebox[\columnwidth][c]{%
            \parbox[c][\subfigheight][c]{\rotlabelwidth}{\centering\rotatebox{90}{\small T-GVC}}%
            \hspace{1pt}%
            \usebox{\subfigboxB}%
        }%
    }

    % ---------- Subfigure 3 ----------
    \sbox{\subfigboxC}{%
        \begin{minipage}{0.96\columnwidth}
            \centering
            \includegraphics[width=\linewidth]{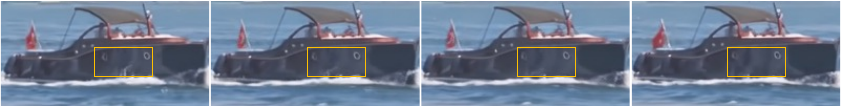}
        \end{minipage}%
    }
    \setlength{\subfigheight}{\dimexpr\ht\subfigboxC+\dp\subfigboxC\relax}

    \subfloat[0.00405 / 4.94 / 0.1587 / 0.0063]{%
        \makebox[\columnwidth][c]{%
            \parbox[c][\subfigheight][c]{\rotlabelwidth}{\centering\rotatebox{90}{\small \textbf{DGVC}}}%
            \hspace{1pt}%
            \usebox{\subfigboxC}%
        }%
    }
    \vspace{-5pt}
    \caption{Visual comparisons with T-GVC on the \textit{Bosphorus} sequence (320$\times$512) from UVG.}
    \label{fig:visualization on Bosphorus}
    \vspace{-15pt}
\end{figure}

\subsubsection{Quantitative Results}
TABLE~\ref{tab:comparison} shows that \ours consistently outperforms T-GVC, establishing state-of-the-art performance within the controllable compression paradigm of generative video compression and demonstrating the advantage of dense, codec-friendly priors over sparse controls. \ours also achieves the best BD-rate with respect to NIQE and t-LPIPS, indicating improved perceptual naturalness and temporal coherence. For DISTS, \ours outperforms all baselines except PLVC and DiffVC, but the two methods may produce unrealistic textures, as illustrated by the PLVC example in Fig.~\ref{fig:visualization on videoSRC26}. For PSNR, \ours substantially outperforms PLVC and DiffVC, and approaches the fidelity-oriented codecs VTM-17.0 and DCVC-FM at low bitrates, as shown in Fig.~\ref{fig:RD curve comparison}. 

\begin{table}[t]
\caption{Per-frame runtime* of T-GVC and \ours on the 320$\times$512 sequence used in the T-GVC paper. Runtime is averaged across bitrate points.}
\centering
\vspace{-5pt}
\renewcommand{\arraystretch}{1.05}
\renewcommand{\tabcolsep}{10pt}
\begin{tabular}{cc|cc}
\toprule
\multicolumn{2}{c|}{Encoding runtime (ms)} & \multicolumn{2}{c}{Decoding runtime (ms)} \\
T-GVC & \ours & T-GVC & \ours \\
\midrule
1614 & 8503 & 878.6 & 584.3 \\
\bottomrule
\end{tabular}
\\
\vspace{2pt}
\footnotesize{*Runtime is tested on NVIDIA H20 and Intel Xeon Platinum 8469C.}
\label{tab:runtime}
\vspace{-6pt}
\end{table}

\begin{table}[t]
\caption{BD-rate (\%)$\downarrow$ / BD-metric$\uparrow$ for ablation studies.}
\centering
\vspace{-4pt}
\renewcommand{\arraystretch}{1.05}
\renewcommand{\tabcolsep}{2pt}
\scalebox{0.95}{
\begin{tabular}{c|c|c|c|c}
\toprule
Method & NIQE & DISTS & t-LPIPS & PSNR(dB) \\ 
\midrule
w/o prior & 133 / -1.14 & 88.9 / -0.0251 & N/A / -0.0371 & N/A / -9.32 \\ 
w/ poses & 41.4 / -0.52 & 35.1 / -0.0062 & N/A / -0.0195 & N/A / -8.87 \\
w/ edge maps & 65.2 / -0.86 & 16.8 / -0.0043 & 94.1 / -0.0038 & N/A / -4.65 \\
w/ sparse trajectories & 28.9 / -0.35 & 28.2 / -0.0058 & N/A / -0.0122 & N/A / -5.91 \\
\midrule
keyframe interval=16 & 19.3 / -0.23 & 5.94 / -0.0018 & 27.1 / -0.0009 & 3.23 / -0.18 \\
keyframe interval=32 & 10.8 / -0.14 & 7.80 / -0.0019 & 5.83 / -0.0002 & 7.91 / -0.23 \\
keyframe interval=96 & -7.74 / 0.05 & 18.5 / -0.0052 & 19.4 / -0.0004 & 19.4 / -0.86 \\
\midrule
w/o color correction & 4.25 / -0.02 & 1.43 / -0.0010 & 0.45 / -0.0001 & 17.7 / -0.73 \\ 
\midrule
\ours & 0.00 / 0.00 & 0.00 / 0.0000 & 0.00 / 0.0000 & 0.00 / 0.00 \\ 
\bottomrule
\end{tabular}
}
\label{tab:ablation}
\vspace{-10pt}
\end{table}

\subsubsection{Qualitative Results}
As illustrated in Fig.~\ref{fig:visualization on videoSRC26} and Fig.~\ref{fig:visualization on Bosphorus},  
compared with VTM-17.0 and DCVC-FM, \ours preserves finer textures and reduces over-smoothing. \ours also avoids unrealistic or unstable detail often produced by PLVC. At similar bitrates, \ours produces more faithful structures and temporally stable textures than T-GVC. These results suggest that dense luminance priors provide stronger spatial constraints for controllable generation than sparse controls.

\subsubsection{Complexity}
Table~\ref{tab:runtime} compares the per-frame runtime of the two controllable compression methods, \textit{i.e.}, T-GVC and \ours. \ours surpasses T-GVC in decoding speed, yet lags in encoding speed due to the high complexity of VVenC. These results suggest that \ours is more suitable for offline applications, while its codec-agnostic design can benefit from future advances in video codecs to reduce encoding time.

\subsection{Ablations}
\label{Ablations}

Table~\ref{tab:ablation} reports ablation results on HEVC B. We compare luminance priors with several sparse alternatives, including poses \cite{dwpose}, edge maps \cite{edge}, and sparse trajectories following T-GVC \cite{wang2026t}, all evaluated with the same generation backbone to isolate the effect of the control signal. Luminance priors consistently outperform these sparse controls, while removing priors causes substantial degradation across all metrics, confirming the importance of dense guidance. Replacing the proposed keyframe selection with uniform intervals of 16, 32, and 96 frames leads to worse results, showing that color-aware keyframe selection provides a better trade-off between bitrate and signal recovery. Removing color correction degrades all metrics, especially PSNR, suggesting that the transmitted frame-level statistics primarily compensate for global color shift and improve pixel-wise fidelity.

\section{Conclusion}
We propose \ours, a dense-prior-guided generative video compression framework that uses luminance components as dense priors to guide reconstruction with a controllable video generation model, without additional training. Experiments show that \ours outperforms prior controllable-compression methods and remains competitive with post-enhancement approaches. These results further demonstrate the potential of controllable compression for generative video compression.

\bibliographystyle{IEEEtran}
\bibliography{vcip2026}

\vspace{12pt}

\end{document}